\documentclass[10pt, a4paper]{article}
\usepackage{float}
\usepackage{silence}
\usepackage[final]{lrec2026} % this is the new style
\usepackage{float}
\usepackage{xcolor}
\usepackage{booktabs}
\usepackage{multirow}
\usepackage{amsmath} 

\usepackage{tikz}
\usetikzlibrary{arrows.meta,positioning}

\title{CV-18 NER: Augmented Common Voice for Named Entity Recognition from Arabic Speech}
\name{Youssef Saidi, Haroun Elleuch, Fethi Bougares}

\address{
ELYADATA, Paris, France \\
\{youssef.saidi, haroun.elleuch, fethi.bougares\}@elyadata.com
}

\abstract{
End-to-end speech Named Entity Recognition (NER) aims to directly extract entities from speech. Prior work has shown that end-to-end (E2E) approaches can outperform cascaded pipelines for English, French, and Chinese, but Arabic remains under-explored due to its morphological complexity, the absence of short vowels, and limited annotated resources. We introduce \textbf{CV-18 NER}, the first publicly available dataset for NER from Arabic speech, created by augmenting the Arabic Common Voice 18 corpus with manual NER annotations following the fine-grained Wojood schema (21 entity types). We benchmark both pipeline systems (ASR + text NER) and E2E models based on Whisper and AraBEST-RQ. E2E systems substantially outperform the best pipeline configuration on the test set, reaching 37.0\% CoER (AraBEST-RQ 300M) and 38.0\% CVER (Whisper-medium).
Further analysis shows that Arabic-specific self-supervised pretraining yields strong ASR performance, while multilingual weak supervision transfers more effectively to joint speech-to-entity learning, and that larger models may be harder to adapt in this low-resource setting. Our dataset and models are publicly released, providing the first open benchmark for end-to-end named entity recognition from Arabic speech \url{https://huggingface.co/datasets/Elyadata/CV18-NER}.
\\ \newline
}

\begin{document}

\maketitleabstract

\section{Introduction}

NER from speech is a challenging and under-explored task, particularly for Arabic, a language characterized by morphological richness and linguistic complexity. Traditional approaches typically follow a cascaded two-stage pipeline: (i) transcribe speech to text using an Automatic Speech Recognition (ASR) system, and (ii) apply a NER model to the resulting transcription. However, this pipeline suffers from error propagation, where transcription errors negatively impact downstream entity recognition. To mitigate this issue, recent studies have explored end-to-end (E2E) approaches that predict named entities directly from speech, bypassing intermediate transcripts. Such models have shown promising results in high-resource languages, with successful systems reported for English \cite{ayache2025whisperner} \cite{Yadav-2020-EndToEndSpeechNER}, French \cite{mdhaffar22_interspeech} \cite{caubriere-etal-2020-named} \cite{szymanski2023ner-yet}, and Chinese \cite{chinese-ner} \cite{Chen-2022-AISHELLNER}. These studies demonstrate the viability of joint speech-to-entity learning, and motivate extending this paradigm to morphologically rich languages such as Arabic, which remain under-explored due to the limited availability of annotated datasets and standardized benchmarks. To the best of our knowledge, only one prior study has addressed semantic extraction from Tunisian Arabic speech, which differs from our focus and setting.

In this paper, we introduce \textbf{CV-18 NER}, the first publicly available Standard Arabic dataset specifically designed for E2E NER from speech. We leverage the Modern Standard Arabic (MSA) subset of Common Voice 18 \cite{ardila2020commonvoice} and manually annotate it with named entities following the Wojood \cite{jarrar2022wojood} annotation schema. In addition, we propose an E2E Arabic speech NER model based on the Whisper architecture \cite{radford2022whisper}. Our approach modifies the Whisper sequence-to-sequence framework to jointly transcribe speech and predict named entities. By enriching transcripts with inline BIO-style entity tags and adapting the tokenizer accordingly, the model learns to associate acoustic cues with semantic labels directly.

Our contributions are summarized as follows:
\begin{itemize}
    \item We present \textbf{CV-18 NER}, the first publicly available dataset for NER from Arabic speech, manually annotated with 21 entity types based on the Wojood tagset.
    \item We develop an end-to-end Whisper-based model that jointly performs speech transcription and NER.
    \item We compare the performance of Whisper models to the Arabic-specific self-supervised learning (SSL) models AraBEST-RQ~\cite{elleuch2026bestrq} in both cascaded and E2E settings.
    \item We report experimental results demonstrating that semantic supervision through entity-aware training improves both entity tagging and transcription quality in Whisper-medium, achieving a 1.3 absolute WER point reduction over its ASR counterpart.
\end{itemize}

\section{Dataset}
\label{sec:Dataset}

This section introduces the \textbf{CV-18 NER} dataset, which is derived from Common Voice 18~\cite{ardila2020commonvoice}, an open-source speech corpus consisting of crowd-sourced read speech recorded by a large number of speakers in diverse acoustic conditions. The Arabic subset of Common Voice 18 contains approximately \textbf{32 hours} of training data, \textbf{12 hours} for development, and \textbf{12 hours} for testing. The corpus is released under a CC0 license, allowing unrestricted redistribution of the annotated data, and has been widely used as a benchmark for Arabic ASR systems.

Common Voice was selected as the basis for CV-18 NER for four main reasons. First, its open license makes it possible to publicly release the resulting annotations. Second, its widespread adoption within the speech processing community makes it a familiar and reproducible benchmark. Third, its moderate size makes large-scale manual annotation feasible while still preserving substantial speaker and recording variability. Last, unlike task-specific speech corpora, the diversity and general-domain nature of Common Voice make it particularly suitable for building realistic speech understanding benchmarks.

\subsection{Named Entity Schema}

For the NER tagset, we conducted a comparative analysis of major Arabic NER corpora, including \textbf{AQMAR} \cite{mohit2012aqmar}, \textbf{ANERCorp} \cite{Benajiba2009ANERsysAA}, and \textbf{Wojood} \cite{jarrar2022wojood}. These corpora differ substantially in tagset granularity. ANERCorp uses a coarse annotation scheme with four entity types (PER, LOC, ORG, MISC), while AQMAR provides a moderately fine-grained annotation with fewer than a dozen categories. In contrast, the \textbf{Wojood} schema defines a richer tagset comprising 21 entity types~\cite{jarrar2022wojood}, covering a broader range of semantic categories such as occupations, nationalities, dates, numerical expressions, and websites.

Wojood was therefore selected for CV-18 NER because its fine-grained schema enables more detailed semantic annotation while remaining compatible with both Modern Standard Arabic and dialectal varieties. This level of granularity is particularly suitable for speech understanding tasks, where diverse entity types frequently occur.

\subsection{Data annotation}
\label{subsec:annotation}

Our data annotation is a pipeline of two sequential stages designed to ensure both annotation efficiency and quality.
\begin{enumerate}
    \item \textbf{Automatic pre-annotation}: In the first stage, several Arabic language models were fine-tuned and evaluated on the Wojood training set. Among the evaluated models, \textbf{AraBERT v2} \cite{antoun2020arabert} achieved the best performance and was therefore selected to generate initial pseudo-labeled named entity annotations for the manual transcriptions of all Common Voice 18 recordings. This pre-annotation step significantly reduced the manual annotation effort while providing an initialization aligned with the Wojood annotation guidelines.
    
    We opted for supervised BERT-based models rather than large language model (LLM) annotation because the task requires strict adherence to the Wojood annotation schema. Fine-tuned NER models produce structured and schema-consistent predictions, whereas prompting-based approaches with LLMs often generate inconsistent label formats and require substantial post-processing.

    \item \textbf{Manual revision}: In the second stage, all pseudo-labeled transcriptions produced during pre-annotation were manually reviewed and corrected
    % \footnote{No inter-annotator agreement score is reported because only one annotator was involved.}.

    This step was performed by a single annotator who corrected the output of the automatic pre-annotation and ensured its compliance with the Wojood annotation schema.   
    Hence, the inter-annotator agreement score is not reported. 
    We acknowledge that relying on a single annotator can introduce bias and aim to mitigate this limitation when additional annotations can be afforded.
    
    % Due to the availability of only a single annotator at the time of the dataset creation, we relied on a hybrid annotation pipeline to mitigate this lack of multiple perspectives. 
    % Thus, no inter-annotator agreement metrics can be 
    
    % This step ensured that the final annotations follow the Wojood schema and remain consistent across the dataset. Since the annotation was performed by a single annotator, inter-annotator agreement measures were not applicable. The use of model-assisted pre-annotation and explicit annotation guidelines helped maintain consistency throughout the corpus.

\end{enumerate}

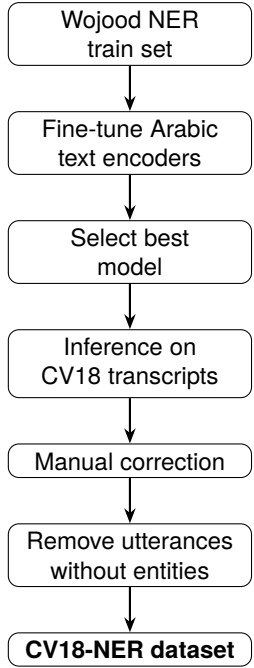
\begin{figure}[t]
    \centering
    \begin{tikzpicture}[
  node distance=6mm,
  box/.style={
    draw,
    rounded corners,
    align=center,
    inner sep=3pt,
    font=\small,
    minimum width=3.2cm
  },
  arrow/.style={-{Stealth[length=2mm]}, thick}
]

\node[box] (wojood) {Wojood NER\\train set};

\node[box, below=of wojood] (ft)
{Fine-tune Arabic\\text encoders};

\node[box, below=of ft] (select)
{Select best\\model};

\node[box, below=of select] (infer)
{Inference on\\CV18 transcripts};

\node[box, below=of infer] (manual)
{Manual correction};

\node[box, below=of manual] (filter)
{Remove utterances\\without entities};

\node[box, below=of filter] (cv18ner)
{\textbf{CV18-NER dataset}};

\draw[arrow] (wojood) -- (ft);
\draw[arrow] (ft) -- (select);
\draw[arrow] (select) -- (infer);
\draw[arrow] (infer) -- (manual);
\draw[arrow] (manual) -- (filter);
\draw[arrow] (filter) -- (cv18ner);

\end{tikzpicture}
    \caption{Annotation workflow of CV-18 NER dataset.}
    \label{fig:annotation_workflow}
\end{figure}

Figure~\ref{fig:annotation_workflow} illustrates the annotation workflow used to construct the CV-18 NER dataset, from model-based pre-annotation to manual revision and filtering.\\

\begin{figure}
    \centering
    \includegraphics[scale=.14]{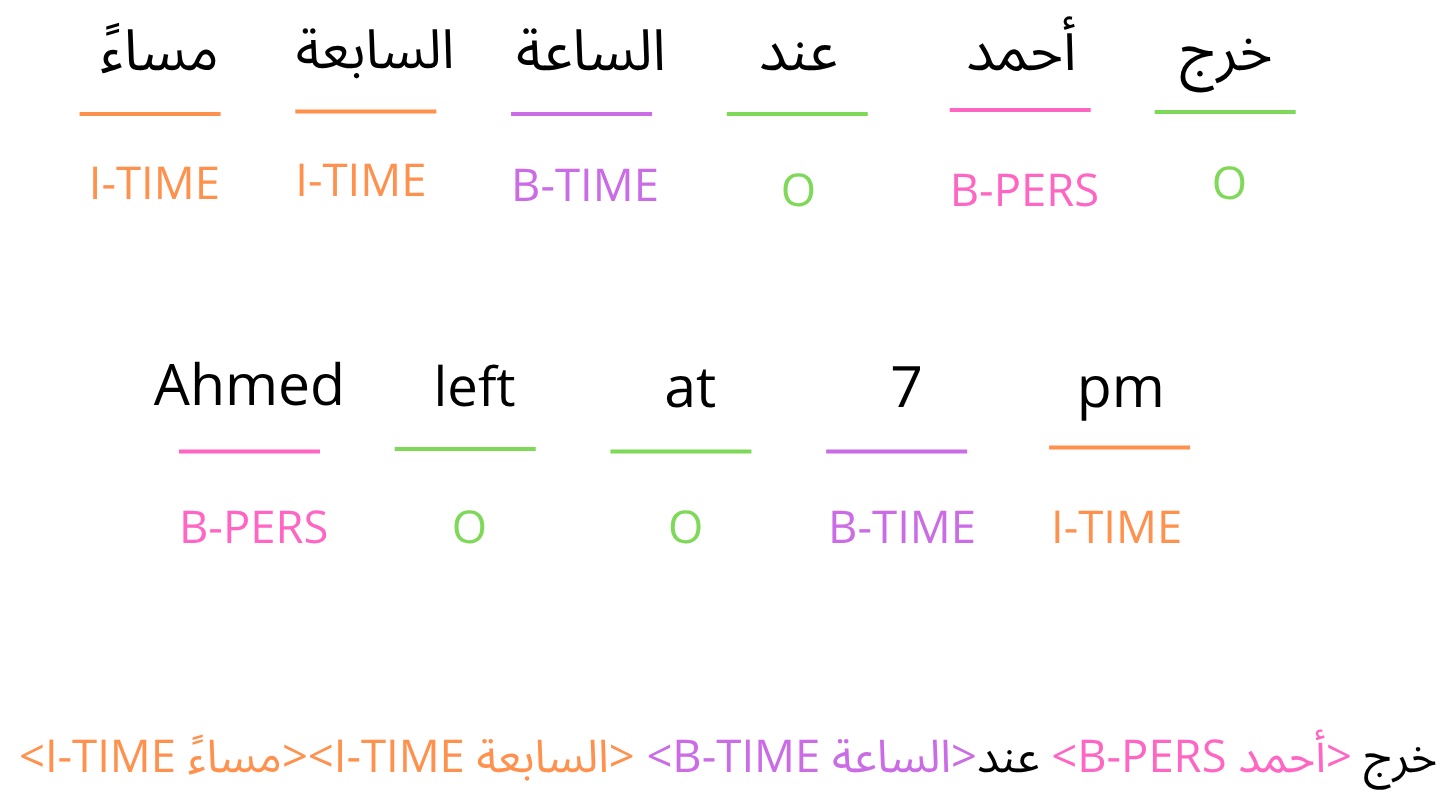}
    \caption{Example of enriched transcripts with inline BIO-style entity markup.}
    \label{fig:sentenceTag}
\end{figure}

The CV-18 NER annotations follow a BIO tagging scheme consistent with the Wojood annotation guidelines. Each entity span is labeled using beginning (B-) and inside (I-) tags associated with the corresponding entity type.
Figure~\ref{fig:sentenceTag} illustrates an example of an Arabic sentence with its tagged representation as well as its English translation.

\subsection{CV-18 NER dataset}

The Common Voice 18 dataset includes 28,410 utterances for training, 10,471 for development, and 10,471 for testing, totaling 32 hours, 12 hours, and 12 hours of speech, respectively.

%The Common Voice 18 dataset has 28,410, 10,471, and 10,471 utterances for the train, dev, and test splits, respectively. This corresponds to 32h, 12h, and 12h of speech, respectively.

Following the annotation pipeline described above and illustrated in Figure~\ref{fig:annotation_workflow}, utterances that do not contain any Named Entity were removed to focus the experiments on informative segments. After applying this filtering step, the number of utterances in each split was reduced by roughly 75\%. We refer to the resulting subset as \textbf{CV-18 NER}. The details of \textbf{CV-18 NER} are reported in Table~\ref{tab:split_stats}.
%After this filtering step, the total number of utterances for each split was reduced by about 75\%.
%We call the resulting subset: \textbf{CV-18 NER} dataset.

\begin{table} [!htb]
  \centering
  \begin{tabular}{lccc}
    \hline
    \textbf{Split} & \textbf{Train}  & \textbf{Dev}  & \textbf{Test}  \\
    \hline
    \# Utterances   & 7119 & 2263 & 2325 \\
    Total Duration (h:mm) & 8:15 & 2:54 & 2:59 \\
    \hline
  \end{tabular}
  \caption{Statistics of CV-18 NER dataset.}
  \label{tab:split_stats}
\end{table}

As shown in Table~\ref{tab:split_stats}, 8h15min of the original 32 training hours remain after filtering out segments without Named Entity annotations. Similarly, 2h54min and 2h59min remain in the development and test sets, respectively. The distribution of Named Entity classes for each split is presented in Table~\ref{tab:tag_distribution}.

% Following the annotation pipeline described above and illustrated in Figure~\ref{fig:annotation_workflow}, utterances that do not contain any named entity labels were removed in order to focus the experiments on informative segments.
% The resulting subset constitutes the \textbf{CV-18 NER} dataset.
% Details of \textbf{CV-18 NER} are reported in Table~\ref{tab:split_stats}.

% \begin{table} [!htb]
%   \centering
%   \begin{tabular}{lccc}
%     \hline
%     \textbf{Split} & \textbf{Train}  & \textbf{Dev}  & \textbf{Test}  \\
%     \hline
%     \# Utterances   & 7119 & 2263 & 2325 \\
%     %# number of words ..???
%     Total Duration (h:mm) & 8:15 & 2:54 & 2:59 \\
%     \hline
%   \end{tabular}
%   \caption{Statistics of CV-18 NER dataset.}
%   \label{tab:split_stats}
% \end{table}

% As shown in Table~\ref{tab:split_stats}, 8h 15min of the original 32 training hours remain after filtering out segments without named entity annotations. Likewise, 2h 54min and 2h 59min remain in the development and test sets, respectively. The distribution of named entity classes for each split is presented in Table~\ref{tab:tag_distribution}.\\

\begin{table}[ht]
  \centering
  \begin{tabular}{lccc}
    \hline
    \textbf{Entity Class} & \textbf{Train} & \textbf{Dev} & \textbf{Test} \\
    \hline
    CARDINAL & 311 & 139 & 138 \\
    CURR     & 2  & 0  & 0   \\
    DATE     & 1176 & 516 & 597 \\
    EVENT    & 112 & 26 & 50 \\
    FAC      & 112 & 46 & 58 \\
    GPE      & 541 & 327 & 288 \\
    LANGUAGE & 237 & 108 & 101 \\
    LAW      & 1   & 3   & 2 \\
    LOC      & 201 & 75 & 97 \\
    MONEY    & 51  & 29 & 29 \\
    NORP     & 1280 & 458 & 378 \\
    OCC      & 588 & 167 & 249 \\
    ORDINAL  & 289 & 103 & 87 \\
    ORG      & 211 & 49 & 117 \\
    PERCENT  & 2   & 2   & 7 \\
    PERS     & 5314 & 1250 & 1373 \\
    PRODUCT  & 40  & 2   & 3 \\
    QUANTITY & 28  & 7   & 17 \\
    TIME     & 539 & 242 & 243 \\
    UNIT     & 0   & 0   & 0 \\
    WEBSITE  & 15  & 6   & 8 \\ \hline
  \end{tabular}
  \caption{Distribution of NER-tags in the CV-18 NER dataset.}
  \label{tab:tag_distribution}
\end{table}

As we can see from this NER-tags distribution, CV-18 NER dataset exhibits a highly imbalanced class distribution, with the \texttt{PERS} category dominating all splits, while several entity types such as \texttt{LAW}, \texttt{CURR}, \texttt{PERCENT}, and \texttt{PRODUCT} are only sparsely represented. 
This imbalance can be explained by the combination of adopting the fine-grained Wojood tagset, which provides broad coverage across multiple semantic domains, and the nature of Common Voice prompts, which consist of short crowd-sourced read sentences that are not specifically curated for named-entity richness. 
As a result, frequently occurring entity types such as \texttt{PERS} and \texttt{GPE} are well represented, whereas rarer semantic categories appear only occasionally. 
Although this imbalance poses challenges for learning low-frequency classes, it reflects realistic conditions in speech applications and Named Entity Recognition tasks, where entity occurrences typically follow long-tailed distributions~\cite{Nemoto_2025, dealvis2024survey-longtail, zevallos2023frequency}. The resulting dataset therefore provides a challenging and representative benchmark for evaluating end-to-end speech NER systems under realistic and limited-resource conditions.

\section{Speech NER systems}
\label{sec:architecture}

To assess the effectiveness of directly predicting named entities from Arabic speech, we implemented and compared two system architectures: (i) a pipeline approach of Automatic Speech Recognition (ASR) followed by a NER system applied to the ASR outputs, and (ii) an end-to-end approach that jointly performs transcription and named entity prediction.

\subsection{Model families}
\label{subsec:arch}

We evaluated two complementary families of speech representation models.
The first family consists of Whisper models~\cite{radford2022whisper}, which are weakly supervised multilingual models trained on large-scale speech–text data and widely used as strong baselines in speech processing tasks. We used the \textbf{Whisper-medium} and \textbf{Whisper-large-v3} pretrained models. \\
The second family consists of AraBEST-RQ models~\cite{elleuch2026bestrq}, a self-supervised learning (SSL) model based on the BEST-RQ framework. Unlike Whisper, AraBEST-RQ models are trained primarily on Arabic audio using self-supervised objectives and do not rely on paired speech–text supervision. We evaluated the 300M and 600M parameter variants.\\
The AraBEST-RQ 300M model is pretrained on approximately 6k hours of crawled Arabic speech. The 600M model is available in two variants: one trained on the same 6k-hour corpus and another trained on approximately 14k hours of Arabic speech obtained by combining crawled data with publicly available datasets, including Common Voice 16.\\
In addition to speech models, we evaluated four text-based NER systems namely \textbf{AraBERT v2}, \textbf{AraBERTv0.2},  \textbf{CAMeLbert-MSA}, and \textbf{CAMeLbert-MIX}.\\
Comparing Whisper and AraBEST-RQ allows us to evaluate NER-from-speech performance across different representation paradigms: large-scale weakly supervised multilingual models on the one hand, and language-specific self-supervised representations on the other. We believe that such a comparison provides meaningful baselines for future research on Arabic speech understanding.

\subsection{Pipeline Speech NER}

The pipeline system follows the classical two-stage approach to speech NER. In the first stage, an ASR module transcribes the input speech signal into text. In the second stage, a text-based NER model assigns named entity labels to the resulting transcription.

ASR systems are implemented using the Whisper and AraBEST-RQ models described in Section~\ref{subsec:arch}. The generated transcriptions are then processed by the text-based NER models to produce the final entity annotations.

Fine-tuning was performed using the CV-18 NER training set as described in Table~\ref{tab:split_stats}. This pipeline configuration serves as a  baseline for comparison with the end-to-end approach described below.

\subsection{End-to-End Speech NER}

In contrast to the pipeline approach, end-to-end (E2E) speech NER systems directly predict named entity annotations from raw speech without relying on an intermediate transcription stage. This formulation enables the joint learning of acoustic and semantic representations and reduces the error propagation typically observed in pipeline systems.\\

Our end-to-end models are implemented using Whisper and AraBEST-RQ architectures described in Section~\ref{subsec:arch}, which are fine-tuned to generate entity-aware transcriptions. Entity-aware training is performed by modifying the reference transcripts to include inline entity tags using BIO-style markup tokens, as described in Section~\ref{subsec:annotation}. The models are trained to generate these enriched transcripts directly from speech, allowing them to learn associations between acoustic patterns and semantic labels while preserving the transcription task.

To support structured entity prediction, the tokenizer vocabularies are extended with entity markers treated as special tokens (e.g., \texttt{<B-PERS>}, \texttt{<I-ORG>}), ensuring that entity tags are handled as atomic units during encoding and decoding.

\section{Experiments and Results}
\label{sec:exp}

\subsection{Experimental Setup}

The text encoder models were fine-tuned on a single NVIDIA Tesla P100 GPU with 16GB of memory. All experiments were conducted using a maximum sequence length of 512 tokens and a batch size of 64. We used the AdamW optimizer with a learning rate of $2\times10^{-5}$, $\epsilon=10^{-8}$, and weight decay of 0.01. The  NER model  was trained using CrossEntropyLoss.\\

For speech experiments, audio signals were resampled to 16 kHz and the reference transcriptions were normalized following \cite{chowdhury21_interspeech}. Specifically, we (a) removed punctuation, (b) removed diacritics, Hamzas, and Maddas, and (c) transliterated Eastern Arabic numerals into Western Arabic numerals. All speech experiments were conducted using the SpeechBrain toolkit \cite{ravanelli2021speechBrain}. Models were trained on an NVIDIA V100 GPU with 32\,GB of VRAM, except for Whisper-large-v3 which was trained on an A100 GPU with 80\,GB of VRAM.\\

\textbf{AraBEST-RQ} models use the same architecture and training hyperparameters for both ASR and E2E NER tasks. Encoder representations are passed through a feedforward classification head followed by a linear projection and softmax layer. Training uses the CTC loss with the AdamW optimizer, with learning rates of $3\times10^{-4}$ for newly initialized layers and $5\times10^{-5}$ for the pretrained encoder. Training is performed with a batch size of 8 and gradient accumulation over 2 steps. BEST-RQ models use a character-level tokenizer. The ASR systems use a vocabulary of 41 symbols, while the NER systems use an extended vocabulary of 79 symbols including entity tags. Each entity tag is treated as a single atomic token.\\

\textbf{Whisper-medium and Whisper-large-v3} models are fine-tuned end-to-end using the negative log-likelihood loss and the AdamW optimizer with a learning rate of $1\times10^{-5}$ and weight decay of 0.01. Training is performed with a batch size of 4 and gradient accumulation over 4 steps. For E2E NER experiments, entity tags are added as special tokens to the Whisper tokenizer and the embedding matrix is resized accordingly.\\

\textbf{Text-based NER models} are evaluated using micro-averaged F1-score. Since text-based systems operate on fixed input sentences, predicted labels can be directly aligned with reference labels, making F1-score an appropriate evaluation metric.\\

For speech-based systems, transcription errors and alignment differences make token-level comparison unreliable. Therefore, speech NER systems are evaluated using error-rate metrics derived from speech recognition and spoken language understanding evaluation. Transcription quality is measured using Word Error Rate (WER), computed after removing special entity tokens from the model outputs. Named entity recognition performance is evaluated using Concept Error Rate (CoER) and Concept-Value Error Rate (CVER), which are adapted from spoken language understanding evaluation. WER, CoER, and CVER follow the standard error-rate formulation:

\[
\text{Error Rate} = \frac{S + D + I}{N},
\]

where $S$, $D$, and $I$ denote substitutions, deletions, and insertions, and $N$ is the number of reference units.\\

CoER evaluates the correctness of entity labels only (e.g., \texttt{B-PERS}, \texttt{I-ORG}), independently of the lexical content. CVER evaluates both entity type and entity span by comparing tag–value pairs extracted from the annotated transcripts. Lower values indicate better performance.

\subsection{Pipeline Speech NER results}

In this section, we present the performance of the pipeline Speech NER system. We used Whisper in zero-shot (ZS) and fine-tuning (FT) settings for the speech-to-text stage. AraBEST-RQ was used only under the fine-tuning (FT) setting. For the text-based NER stage, we evaluated the four different models mentioned in \ref{subsec:arch}. Before evaluating the full pipeline (ASR+NER), we first assess each component individually.

\subsubsection{Speech Recognition Results}

We compared AraBEST-RQ to Whisper-medium and Whisper-large-v3 in both zero-shot and fine-tuned settings. Fine-tuning was performed on the filtered training subset (8 hours 15 minutes, Table~\ref{tab:split_stats}). Table~\ref{tab:asr-results} reports the WER obtained by all ASR models on the validation and test splits.

The results show that zero-shot Whisper performs very poorly on CV-18 NER, with WER above 92\% for Whisper-large-v3 and 107\% for Whisper-medium. Fine-tuning substantially improves performance, especially for Whisper-medium, which achieves the best WER among Whisper models with \textbf{14.9\%} on the validation set and \textbf{22.0\%} on the test set. In contrast, Whisper-large-v3 shows a smaller improvement after fine-tuning, reaching 31.8\% on validation and 32.3\% on test, and remains significantly worse than the medium model. Preliminary analysis of its outputs indicates that Whisper-large-v3 sometimes produces English translations instead of Arabic transcriptions or transliterations into Latin script, which warrants further investigation.
Such behaviour is, however, known and has been previously observed and reported in works such as ~\citep{talafha2024casablanca}.

Among AraBEST-RQ models, the 300M 6k variant achieves the lowest validation WER (\textbf{11.2\%}) and the lowest test WER (\textbf{15.1\%}), outperforming all Whisper models. Other AraBEST-RQ variants show slightly higher WERs, with the 600M 6k and 600M 14k models achieving up to 17.4\% on the test set.

Overall, these results highlight that fine-tuning is critical for Whisper-based ASR on limited Arabic data and that increasing model size alone does not guarantee better performance. AraBEST-RQ demonstrates a more consistent improvement across both validation and test sets, making it the preferred choice in this low-resource scenario.
% \begin{table}[!htb]
%     \centering
%     \begin{tabular}{@{}lll@{}}
%     \toprule
%     \multicolumn{1}{c}{\textbf{Model}} & \textbf{Valid.} & \multicolumn{1}{c}{\textbf{Test}} \\ \midrule
%     Whisper-medium (ZS)   & 103.11 & 109.68 \\
%     Whisper-medium (FT)   & \textbf{13.79} & \textbf{20.65} \\
%     Whisper-large-v3 (ZS) & 88.28  & 91.83  \\
%     Whisper-large-v3 (FT) & 30.87  & 31.41  \\ \midrule
%     AraBEST-RQ 300M 6k (FT)  & \textbf{10.16} & \textbf{57.68} \\
%     AraBEST-RQ 600M 6k (FT)  & 11.20 & 58.80 \\ 
%     AraBEST-RQ 600M 14k (FT) & 11.20 & 58.18 \\ 
%     \bottomrule
%     \end{tabular}
%     \caption{Validation and test Word Error Rates (WER \%) of ASR models on the CV-18 NER dataset. FT: Fine-tuning. ZS: Zero-shot.}
%     \label{tab:asr-results}
% \end{table}

\begin{table}[!htb]
    \centering
    \begin{tabular}{@{}lll@{}}
    \toprule
    \multicolumn{1}{c}{\textbf{Model}} & \textbf{Valid.} & \multicolumn{1}{c}{\textbf{Test}} \\ \midrule
    Whisper-medium (ZS)   & 107.4 & 112.3 \\
    Whisper-medium (FT)   & 14.9 & 22.0 \\
    Whisper-large-v3 (ZS) & 92.7  & 95.1  \\
    Whisper-large-v3 (FT) & 31.8  & 32.3  \\ \midrule
    AraBEST-RQ 300M 6k (FT)  & \textbf{11.2} & \textbf{15.1} \\
    AraBEST-RQ 600M 6k (FT)  & 12.2 & 17.4 \\ 
    AraBEST-RQ 600M 14k (FT) & \textbf{11.2} & 16.4 \\ 
    \bottomrule
    \end{tabular}
    \caption{Validation and test Word Error Rates (WER \%) of ASR models on the CV-18 NER dataset. FT: Fine-tuning. ZS: Zero-shot. Best scores per metric are reported in bold.}
    \label{tab:asr-results}
\end{table}
    
% As shown in Table \ref{tab:asr-results}, the WER decreases from xx\% to yy\% and from 26.53\% to zz\% on dev and test sets, respectively. 
% As expected, results are better with Whisper large-v3. Under zero-shot setting, Whisper large-v3 gives a WER of 12.59\% and 20.97\% on dev and test sets, respectively. Word Error Rate further decreases when Whisper large-v3 is finetuned to achieve 10.46\% and 14.59\% WER on dev and test sets, respectively.

\subsubsection{Text-based NER Results}
We fine-tune several Arabic BERT-based~\cite{devlin2019bert} models on the CV-18 NER train set transcriptions and evaluate them using \textbf{micro-F1} on the transcriptions of the validation and test sets of the same dataset. The top  performing models were then used for pipeline systems.

\begin{table}[!htb]
    \centering
    \begin{tabular}{@{}lll@{}}
    \toprule
    \multicolumn{1}{c}{\textbf{Model}} & \textbf{Validation} & \multicolumn{1}{c}{\textbf{Test}} \\ \midrule
    CamelBERT-Mix  & 79.66 & 77.24 \\
    CamelBERT-MSA  & 79.76 & 76.13 \\
    AraBERTv2      & 81.06 & 76.58 \\
    AraBERTv0.2    & \textbf{81.10} & \textbf{77.34} \\
    \bottomrule
    \end{tabular}
    \caption{Validation and test micro-F1 scores (\%) of BERT-based NER models on the CV-18 NER dataset. Best scores per column are reported in bold.}
    \label{tab:text-ner-results}
\end{table}

\subsubsection{Pipeline Systems Results}

Based on the ASR and text NER results reported above, we evaluated the Speech NER pipeline system using the five fine-tuned ASR systems, followed by the four chosen BERT-based systems to perform NER on the generated transcriptions. 
The obtained results are presented in Table~\ref{tab:pipeline-test-compact}.\\

While transcription quality generally influences downstream performance, the results show that lower WER does not necessarily lead to better entity extraction. In particular, Whisper-medium pipelines outperform AraBEST-based pipelines despite higher transcription error rates. \\

On the CV-18 NER test split, the lowest CoER is obtained when AraBERTv0.2 is applied to the output of the fine-tuned  Whisper-medium ASR system, achieving 51.3 CoER and 50.2 CVER. These results suggest that while ASR quality influences downstream entity extraction, it is not the only factor; the choice of ASR and NER models (and the nature of ASR errors) substantially impacts the final pipeline performance.\\

%\begin{table}[!htb]
 %   \centering
  %  \scriptsize
   % \begin{tabular}{@{}l l cc@{}}
    %\toprule
    %\textbf{NER Model} & \textbf{ASR Model} & %\textbf{CoER} & \textbf{CVER} \\ \midrule
    
    %\multirow{5}{*}{CamelBERT-Mix} 
     %   & AraBEST-RQ 600M 14k       & 41.8318 & 61.3552 \\
      %  & AraBEST-RQ 300M 6k    & 41.803  & 60.8384 \\
     %   & AraBEST-RQ 600M 6k    & 42.6357 & 61.9294 \\
      %  & Whisper-large-v3 (FT) & 49.0669 & 64.255  \\
       % & Whisper-medium (FT) & 42.0902 & 61.9007 \\ \midrule

   % \multirow{5}{*}{CamelBERT-MSA} 
    %    & AraBEST-RQ 600M 14k       & 42.5782 & 61.3839 \\
     %   & AraBEST-RQ 300M 6k    & 42.1476 & 60.4077 \\
      %  & AraBEST-RQ 600M 6k    & 42.9515 & 61.4987 \\
       % & Whisper-large-v3 (FT) & 49.2966 & 64.1688 \\
        %& Whisper-medium (FT) & 42.5782 & 61.8145 \\ \midrule

    %\multirow{5}{*}{AraBERTv2} 
     %   & AraBEST-RQ 600M 14k      & 48.5501 & 71.3753 \\
      %%  & AraBEST-RQ 300M 6k    & 48.9808 & 71.4327 \\
        %& AraBEST-RQ 600M 6k    & 49.2392 & 71.5762 \\
        %& Whisper-large-v3 (FT) & 53.9765 & 71.8346 \\
        %& Whisper-medium (FT) & 48.3778 & 71.1456 \\ \midrule

    %\multirow{5}{*}{AraBERTv0.2} 
     %   & AraBEST-RQ 600M 14k       & 36.9222 & 53.7468 \\
      %  & AraBEST-RQ 300M 6k     & 36.549  & 53.9477 \\
       % & AraBEST-RQ 600M 6k    & 37.1232 & 54.6655 \\
        %& Whisper-large-v3 (FT) & 44.8177 & 58.9147 \\
        %& Whisper-medium (FT) & 37.238  & 55.211 \\
        
    %\bottomrule
    %\end{tabular}
    %\caption{Pipeline evaluation on CV-18 NER dev set with grouped NER models. Columns show CoER and CVER for each ASR + NER combination.}
    %\label{tab:pipeline-dev-compact}
%\end{table}

\begin{table}[!htb]
    \centering
    \scriptsize
    \begin{tabular}{@{}l l cc@{}}
    \toprule
    \textbf{NER Model} & \textbf{ASR Model} & \textbf{CoER} & \textbf{CVER} \\ \midrule
    
    \multirow{5}{*}{CamelBERT-Mix} 
        & AraBEST-RQ 600M 14k        & 66.4 & 65.8 \\
        & AraBEST-RQ 300M 6k        & 66.1 & 65.4 \\
        & AraBEST-RQ 600M 6k        & 67.7 & 67.0 \\
        & Whisper-large-v3 (FT)  & 57.7 & 56.7 \\
        & Whisper-medium (FT) & 57.2 & 55.9 \\ \midrule

    \multirow{5}{*}{CamelBERT-MSA} 
        & AraBEST-RQ 600M 14k        & 66.3 & 65.7 \\
        & AraBEST-RQ 300M 6k        & 66.3 & 65.7 \\
        & AraBEST-RQ 600M 6k        & 67.9  & 67.2 \\
        & Whisper-large-v3 (FT)  & 57.8 & 56.7 \\
        & Whisper-medium (FT) & 57.4 & 56.1 \\ \midrule

    \multirow{5}{*}{AraBERTv2} 
        & AraBEST-RQ 600M 14k         & 72.0 &  71.3 \\
        & AraBEST-RQ 300M 6k        & 71.5 & 70.9 \\
        & AraBEST-RQ 600M 6k       & 72.0 & 71.4 \\
        & Whisper-large-v3 (FT)  & 63.6 & 62.5 \\
        & Whisper-medium (FT) & 64.5 & 63.0 \\ \midrule

    \multirow{5}{*}{AraBERTv0.2$^*$} 
        & AraBEST-RQ 600M 14k         & 63.1 & 62.6 \\
        & AraBEST-RQ 300M 6k        & 62.7 & 62.0 \\
        & AraBEST-RQ 600M 6k        & 63.8  & 63.2 \\
        & Whisper-large-v3 (FT)  & 54.1 & 53.5 \\
        & Whisper-medium (FT)$^*$ & \textbf{51.3} & \textbf{50.2} \\
        
    \bottomrule
    \end{tabular}
    \caption{Pipeline evaluation on CV-18 NER test set with grouped NER models. Columns show CoER and CVER for each ASR + NER combination. Best scores per metric are reported in bold.  FT: Fine-tuning. The best configurations are followed by $^*$.}
    \label{tab:pipeline-test-compact}
\end{table}
A qualitative analysis shows that the best pipeline system performs well on high-frequency entity types such as \texttt{\textbf{LANGUAGE}} and \texttt{\textbf{GPE}}, while performance drops sharply for low-frequency categories such as \texttt{\textbf{PRODUCT}}, \texttt{\textbf{LAW}}, and \texttt{\textbf{QUANTITY}}.
We also noted that ASR errors, particularly those stemming from phonetic ambiguities in MSA, contribute significantly to the degradation of NER performance.

\subsection{End-to-end Speech NER results}

This section presents the performance of our E2E Speech NER system, which is based on Whisper and AraBEST-RQ models fine-tuned in an E2E fashion for entity extraction.

Table~\ref{tab:e2e-results} shows that the AraBEST-RQ 600M models failed to produce correct outputs in E2E fine-tuning, leading to near-random predictions. However, AraBEST-RQ 300M outperforms all other systems in WER and CoER. Whisper-medium is, however, the best in terms of CVER. Whisper-large-v3 is also largely outperformed by the smaller Whisper-medium and AraBEST-RQ 300M.
% \textcolor{red}{Detailed per-entity results for the two best-performing models, Whisper-medium and AraBEST-RQ 300M 6k, are provided in the Appendix.}

\begin{table}[!htb]
    \centering
    \begin{tabular}{@{}llll@{}}
    \toprule
    \multicolumn{1}{c}{\textbf{Model}} & \multicolumn{1}{c}{\textbf{WER}} & \multicolumn{1}{c}{\textbf{CoER}} & \multicolumn{1}{c}{\textbf{CVER}} \\ \midrule
    AraBEST-RQ 300M 6k   & \textbf{16.0} & \textbf{37.0}          & 43.5          \\
    AraBEST-RQ 600M 6k   & 99.4          & 87.1          & 94.0          \\ 
    AraBEST-RQ 600M 14k  & 99.6          & 99.8          & 100              \\ \midrule
    Whisper-medium      & 20.7           & 38.8 & \textbf{38.0} \\
    Whisper-large-v3   & 41.9          & 59.6          & 59.1          \\ \bottomrule
    \end{tabular}
    \caption{E2E NER from speech results on the CV-18 NER test set. Best scores per metric are reported in bold. Entity tags are removed before WER computation. The same normalization is applied to all systems.}
    \label{tab:e2e-results}

\end{table}

We did the same entity-wise investigation, and the results show improved performance on several core entity types. 
Notably, the E2E model achieves better results for \texttt{\textbf{LANGUAGE}} and \texttt{\textbf{ORDINAL}}, along with significant gains on \texttt{\textbf{PERS}} and \texttt{\textbf{GPE}} compared to the pipeline.
However, performance remains limited on low-frequency types such as \texttt{\textbf{LAW}}, \texttt{\textbf{PERCENT}}, and \texttt{\textbf{PRODUCT}}, which were completely missed or poorly predicted.
These results demonstrate the potential of end-to-end training to capture richer acoustic-entity correlations, while also highlighting ongoing challenges in recognizing rare entity types from speech. A detailed breakdown of errors per entity for the two best-performing models, Whisper-medium and AraBEST-RQ 300M 6k, can be found in Tables~\ref{tab:whisper_entity} and~\ref{tab:bestrq_entity} in the Appendix.

\section{Discussion and Analysis}

This section provides a detailed analysis of the experimental findings across both pipeline and end-to-end (E2E) configurations, with particular emphasis on representation paradigms, error propagation, and entity-type behavior.

\subsection{ASR Quality vs. Downstream Entity Recognition}

Despite AraBEST-RQ 300M achieving substantially lower WER than Whisper-medium, Whisper-based pipelines consistently obtain lower CoER and CVER scores. This suggests that transcription quality alone does not fully determine entity extraction performance.

These findings confirm that classical pipeline systems remain highly sensitive to transcription errors, particularly for entity spans where even a single token substitution may invalidate the entire tag–value pair, resulting in high CVER.

\subsection{Weakly Supervised vs. Self-Supervised Representations}

A central contribution of this work is the comparison between Whisper (weakly supervised multilingual training) and AraBEST-RQ (Arabic-specific self-supervised learning).\\

In the pipeline setting, AraBEST-RQ 300M performs strongly as an ASR model, demonstrating that Arabic-specific SSL representations are competitive when fine-tuned for transcription, with all AraBEST-RQ models outperforming Whisper models. However, this pattern shifts in E2E NER from speech. While AraBEST-RQ 300M is still the best-performing model in both WER and CoER, it falls behind Whisper-medium by 5.5 absolute points in CVER.
It is worth noting that AraBEST-RQ 300M is less than half the size of Whisper-medium (769M parameters) in terms of the number of parameters.

This contrast suggests that:
\begin{itemize}
    \item SSL-based encoders trained without paired speech-text supervision adapt well to ASR through CTC fine-tuning,
    \item but larger SSL-based models struggle to converge in joint transcription and semantic tagging, as observed for the AraBEST-RQ 600M variants.\\
\end{itemize}

% In contrast, AraBEST-RQ 300M achieves the best overall CoER (37.0\%), while Whisper-medium obtains the best CVER (38.0\%), demonstrating competitive end-to-end NER performance. 
We also observe that well-tuned smaller-scale models can provide strong alignment between acoustic and lexical-semantic representations, which is crucial for direct entity prediction. Interestingly, Whisper-large-v3 does not outperform Whisper-medium in E2E training. This may be attributed to over-parameterization relative to the limited 8-hour training subset, leading to suboptimal adaptation under low-resource fine-tuning conditions. This finding aligns with recent observations in Arabic speech processing~\cite{elleuch2025elyadatalianadi2025asr}, where Whisper-medium similarly outperformed Whisper-large-v3 when fine-tuned on dialect-specific data of a similar scale (~16 hours) for ASR tasks.

\subsection{End-to-End vs. Pipeline: Error Propagation Revisited}
 
Comparing the best pipeline configuration (AraBERTv0.2 + Whisper-medium ) with the best E2E results, we observe:

\begin{itemize}
    \item Lower CoER in E2E (37.0 vs. 51.3),
    \item Lower CVER in E2E (38.0 vs. 50.2).
\end{itemize}

These improvements confirm that eliminating the intermediate transcription stage reduces error propagation. 
In pipeline systems, entity prediction depends strictly on lexical correctness. 
Any ASR substitution inside an entity span typically results in a CVER penalty.\\

In contrast, the E2E model can partially recover entity type information from acoustic cues even when lexical realization is imperfect. 
This indicates that semantic supervision encourages the model to encode entity-relevant acoustic patterns, rather than relying exclusively on textual reconstruction. Additionally, E2E training appears to act as a form of task-aware regularization, improving semantic robustness even when transcription accuracy (WER) is not the lowest among models.

% Figure~\ref{fig:pipeline-vs-e2e-error} shows an example of outputs from the best performing E2E and pipeline system. It shows a case of ASR error leading to a NER classification error, which is circumvented in E2E settings.

% \begin{figure}[!htb]
%     \centering
%     \includegraphics[width=0.57\textwidth]{pipeline vs E2E error.png}
%     \caption{Example of error differences between pipeline and end-to-end NER systems. These examples are taken from the best performing pipeline and E2E systems.}
%     \label{fig:pipeline-vs-e2e-error}
% \end{figure}

\subsection{Entity-Type Behavior }

The dataset has an imbalanced distribution, with \texttt{PERS}, \texttt{GPE}, and \texttt{DATE} dominating the corpus, while categories such as \texttt{LAW}, \texttt{PRODUCT}, and \texttt{PERCENT} are extremely sparse.

Both pipeline and E2E systems show:

\begin{itemize}
    \item Strong performance on high-frequency classes (\texttt{PERS}, \texttt{GPE}, \texttt{LANGUAGE}),
    \item Severe degradation on rare classes (\texttt{LAW}, \texttt{PRODUCT}, \texttt{QUANTITY}).
\end{itemize}

However, the E2E model demonstrates improved robustness on certain mid-frequency categories such as \texttt{ORDINAL} and \texttt{TIME}, suggesting that joint acoustic-semantic training may help disambiguate numeric and temporal expressions that are phonetically structured.

Rare classes remain challenging for both paradigms. 
Given the extremely small number of training examples (e.g., \texttt{LAW} and \texttt{CURR}), the models are unable to learn stable acoustic-semantic mappings. 

\subsection{Key Takeaways}

From these experiments, we draw several conclusions:

\begin{enumerate}
    \item End-to-end training substantially reduces entity-level error compared to strong pipeline baselines.
    \item Language-specific SSL representations can be competitive for E2E NER from speech with a 300M parameters model outperforming Whisper-medium in both WER and CoER despite being less than half its size.
    \item Smaller models show better performance in ASR and E2E NER from speech than their larger counterparts.
    \item Pipeline systems remain highly sensitive to ASR errors, especially at the entity-span level (CVER).
    \item Entity imbalance remains the primary bottleneck for Arabic speech NER.
\end{enumerate}

Overall, the results demonstrate that E2E speech NER is not only viable for Modern Standard Arabic but can significantly outperform cascade approaches when appropriate pretrained representations are used.

\section{Conclusion}

In this paper, we introduced \textbf{CV-18 NER}, the first annotated Arabic dataset for Speech Named Entity Recognition. Initial experiments with SoTA models demonstrate that  E2E speech NER is viable for Arabic and significantly outperforms cascade approaches when appropriate pretrained representations are used. We believe that this resource, along with the baseline results provided in this paper, will foster the development of better end-to-end speech NER for the Arabic language and its dialects.

\section{Bibliographical References}\label{sec:reference}
\bibliographystyle{lrec2026-natbib}
\bibliography{lrec2026-example}

\newpage

\section*{Appendix}
\appendix
\section{Per-Entity Error Statistics for the Best-Performing E2E Systems}
% Tables~\ref{tab:whisper_entity} and~\ref{tab:bestrq_entity} report the detailed per-entity results for the two best-performing models in the End-to-End (E2E) configuration: Whisper-medium and AraBEST-RQ 300M 6k. 
Tables~\ref{tab:whisper_entity} and~\ref{tab:bestrq_entity} present a breakdown by entity of error types for the two best-performing end-to-end NER-from-speech systems : Whisper-medium and AraBEST-RQ 300M 6k. For each entity tag, we report the number of insertions (INS), deletions (DEL), and substitutions (SUB), as well as the Concept Error Rate (CoER, \%).

\begin{table}[ht]
\centering
\caption{Per-entity CoER results for Whisper-medium (E2E configuration). INS: insertions, DEL: deletions, SUB: substitutions.}
\label{tab:whisper_entity}
\resizebox{\columnwidth}{!}{%
\begin{tabular}{lcccc}
\hline
\multicolumn{1}{c}{\textbf{Tag}} & \textbf{INS} & \textbf{DEL} & \textbf{SUB} & \textbf{CoER (\%)} \\
\hline
B-CARDINAL  & 2  & 10  & 31 & 36.8  \\
I-CARDINAL  & 1  & 4   & 9  & 93.3  \\
B-DATE      & 29 & 43  & 42 & 31.2  \\
I-DATE      & 11 & 59  & 46 & 50.4  \\
B-EVENT     & 4  & 14  & 10 & 112.0 \\
I-EVENT     & 4  & 8   & 14 & 108.3 \\
B-FAC       & 1  & 19  & 23 & 95.6  \\
I-FAC       & 1  & 2   & 8  & 100.0 \\
B-GPE       & 11 & 27  & 74 & 42.3  \\
I-GPE       & 2  & 5   & 11 & 81.8  \\
B-LANGUAGE  & 0  & 7   & 4  & 10.9  \\
B-LAW       & 0  & 1   & 0  & 100.0 \\
B-LOC       & 16 & 26  & 23 & 108.3 \\
I-LOC       & 0  & 14  & 23 & 100.0 \\
B-MONEY     & 0  & 5   & 5  & 76.9  \\
I-MONEY     & 1  & 3   & 8  & 75.0  \\
B-NORP      & 16 & 72  & 53 & 49.0  \\
I-NORP      & 7  & 30  & 33 & 77.8  \\
B-OCC       & 10 & 42  & 45 & 47.1  \\
I-OCC       & 6  & 21  & 18 & 104.7 \\
B-ORDINAL   & 2  & 7   & 24 & 38.4  \\
I-ORDINAL   & 0  & 0   & 1  & 100.0 \\
B-ORG       & 3  & 27  & 14 & 68.8  \\
I-ORG       & 1  & 16  & 24 & 91.1  \\
B-PERCENT   & 0  & 1   & 1  & 100.0 \\
I-PERCENT   & 0  & 0   & 5  & 100.0 \\
B-PERS      & 40 & 125 & 88 & 23.1  \\
I-PERS      & 29 & 61  & 32 & 45.9  \\
B-PRODUCT   & 0  & 1   & 2  & 100.0 \\
I-PRODUCT   & 0  & 0   & 0  & 0.0   \\
B-QUANTITY  & 1  & 3   & 6  & 111.1 \\
I-QUANTITY  & 0  & 1   & 7  & 100.0 \\
B-TIME      & 10 & 26  & 23 & 39.9  \\
I-TIME      & 7  & 20  & 25 & 54.7  \\
B-WEBSITE   & 0  & 0   & 7  & 100.0 \\
I-WEBSITE   & 0  & 0   & 0  & 0.0   \\
\hline
\end{tabular}%
}  
\end{table}

\newpage
\begin{table}[!htb]
\centering
\caption{Per-entity CoER results for AraBEST-RQ 300M 6k (E2E configuration). INS: insertions, DEL: deletions, SUB: substitutions.}
\label{tab:bestrq_entity}
\resizebox{\columnwidth}{!}{%
\begin{tabular}{lcccc}
\hline
\multicolumn{1}{c}{\textbf{Tag}} & \textbf{INS} & \textbf{DEL} & \textbf{SUB} & \textbf{CoER (\%)} \\
\hline
B-CARDINAL  & 4  & 3   & 26 & 28.2  \\
I-CARDINAL  & 0  & 5   & 3  & 53.3  \\
B-DATE      & 34 & 31  & 41 & 29.0  \\
I-DATE      & 17 & 43  & 38 & 42.6  \\
B-EVENT     & 8  & 6   & 7  & 84.0  \\
I-EVENT     & 5  & 9   & 7  & 84.0  \\
B-FAC       & 3  & 21  & 11 & 77.8  \\
I-FAC       & 2  & 4   & 8  & 107.7 \\
B-GPE       & 13 & 24  & 45 & 30.8  \\
I-GPE       & 2  & 3   & 7  & 54.5  \\
B-LANGUAGE  & 1  & 3   & 8  & 11.9  \\
B-LAW       & 0  & 1   & 0  & 100.0 \\
B-LOC       & 11 & 18  & 14 & 71.7  \\
I-LOC       & 2  & 8   & 14 & 64.9  \\
B-MONEY     & 0  & 6   & 0  & 46.2  \\
I-MONEY     & 1  & 0   & 8  & 56.3  \\
B-NORP      & 25 & 57  & 52 & 46.5  \\
I-NORP      & 7  & 31  & 29 & 74.4  \\
B-OCC       & 7  & 38  & 23 & 33.0  \\
I-OCC       & 5  & 9   & 14 & 65.1  \\
B-ORDINAL   & 10 & 6   & 15 & 36.0  \\
I-ORDINAL   & 0  & 0   & 0  & 0.0   \\
B-ORG       & 5  & 17  & 7  & 44.6  \\
I-ORG       & 4  & 12  & 12 & 56.0  \\
B-PERCENT   & 0  & 0   & 2  & 100.0 \\
I-PERCENT   & 0  & 0   & 5  & 100.0 \\
B-PERS      & 70 & 86  & 39 & 17.7  \\
I-PERS      & 37 & 52  & 29 & 43.9  \\
B-PRODUCT   & 0  & 1   & 2  & 100.0 \\
I-PRODUCT   & 0  & 0   & 0  & 0.0   \\
B-QUANTITY  & 1  & 1   & 3  & 55.6  \\
I-QUANTITY  & 4  & 2   & 1  & 87.5  \\
B-TIME      & 7  & 12  & 21 & 27.0  \\
I-TIME      & 6  & 22  & 11 & 41.1  \\
B-WEBSITE   & 1  & 2   & 2  & 62.5  \\
I-WEBSITE   & 0  & 0   & 0  & 0.0   \\
\hline
\end{tabular}%
}
\end{table}
\end{document}